\documentclass{article}
\usepackage{ijcai24}

\usepackage{times}
\usepackage{soul}
\usepackage{url}
\usepackage[hidelinks]{hyperref}
\usepackage[utf8]{inputenc}
\usepackage[small]{caption}
\usepackage{graphicx}
\usepackage{amsmath}
\usepackage{amsthm}
\usepackage{booktabs}
\usepackage{algorithm}
\usepackage{algorithmic}
\usepackage[switch]{lineno}
\graphicspath{{figures/}} 
\graphicspath{{figures/}} 
\newcommand{\ours}{IE-NeRF} 
\usepackage{subfig}
\usepackage{amsmath}
\usepackage{amsthm}
\usepackage{amssymb,amsfonts}
\usepackage{latexsym}
\usepackage[commandnameprefix=always, final, markup=default, authormarkup=none]{changes}

\title{IE-NeRF: Inpainting Enhanced Neural Radiance Fields in the Wild}

\iftrue
\author{%
 Shuaixian~Wang$^1$,~
Haoran~Xu$^{1,2}$,~
Yaokun~Li$^1$,~
Jiwei~Chen$^1$,~
Guang~Tan$^{1*}$\\
  \normalsize $^1$Sun Yat-sen University, Guangdong, China\\
$^2$Pengcheng Laboratory, Shenzhen, China\\
  {\{wangshx29, xuhr9, liyk58, chenjw269\}@mail2.sysu.edu.cn,
tanguang@mail.sysu.edu.cn} \\
}
\fi

\begin{document}

\maketitle

\begin{abstract}
   We present a novel approach for synthesizing realistic novel views using Neural Radiance Fields (NeRF) with uncontrolled photos in the wild. While NeRF has shown impressive results in controlled settings, it struggles with transient objects commonly found in dynamic and time-varying scenes. Our framework called \textit{Inpainting Enhanced NeRF}, or \ours, enhances the conventional NeRF by drawing inspiration from the technique of image inpainting. Specifically, our approach extends the Multi-Layer Perceptrons (MLP) of NeRF, enabling it to simultaneously generate intrinsic properties (static color, density) and extrinsic transient masks. We introduce an inpainting module that leverages the transient masks to effectively exclude occlusions, resulting in improved volume rendering quality. Additionally, we propose a new training strategy with frequency regularization to address the sparsity issue of low-frequency transient components. We evaluate our approach on internet photo collections of landmarks, demonstrating its ability to generate high-quality novel views and achieve state-of-the-art performance.
\end{abstract}

\section{Introduction}\label{sec:intro}
Synthesizing novel views of a scene from limited captured images is a long-standing problem in computer vision, which is fundamental for applications in mixed reality~\cite{vr_nerf}, 3D reconstruction~\cite{3drec,zhou2024neural}. Canonical view synthesizing techniques~\cite{knorr2007image,szeliski2022image} based on structure-from-motion and image rendering have encountered challenges in maintaining consistency across views, as well as addressing occlusion and distortion. Recently, with the development of implicit scene representation and neural rendering, Neural Radiance Fields (NeRF)~\cite{nerf} have achieved excellent performance in novel view synthesis (NVS). 


NeRF employs neural networks to encode radiance properties within a continuous spatial domain, enabling intricate scene reconstructions by learning from multiple viewpoints. While this approach has achieved success in various fields such as computer graphics, computer vision, and immersive technologies~\cite{nerf_sota}, conventional NeRFs often operate under controlled settings with static scenes and consistent lighting conditions~\cite{wild19}. However, in real-world scenarios characterized by time-varying and transient occlusions, NeRF encounters significant performance degradation.
\begin{figure}[]
\centering
\newcommand{\CCC}{0.70\linewidth}
\subfloat{
    \includegraphics[width=\CCC]{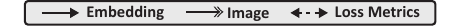}
}\\
\setcounter{subfigure}{0}
\subfloat[Dual NeRFs]{
    \includegraphics[width=\CCC]{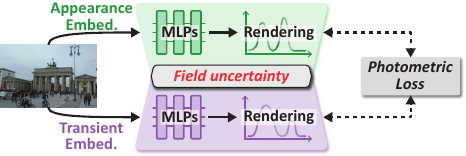}
    \label{fig:moti:a}
}\\
\subfloat[Prior-Assisted NeRFs]{
    \includegraphics[width=\CCC]{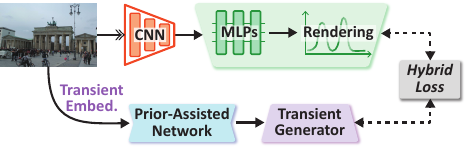}
    \label{fig:moti:b}
}\\
\subfloat[\ours\ (Ours)]{
    \includegraphics[width=\CCC]{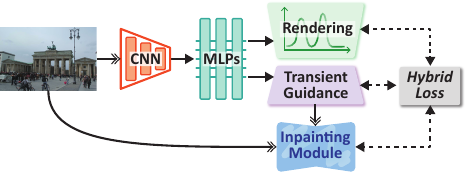}
    \label{fig:moti:c}
}
\caption{
    Comparison of different NeRF pipelines that mitigate transient occlusion.
    (a) Dual NeRFs, which extract transient components by introducing an additional NeRF branch; 
    (b) Prior-Assisted NeRFs, which leverage prior knowledge to assist in separating transient objects; 
    (c) \ours\ (Ours), which integrates the inpainting module to enhance NeRF. 
}
\label{fig:motivation}
\end{figure}
Existing solutions to this problem can be roughly categorized into two approaches
\textbf{(i) Dual NeRFs}. As depicted in \figurename~\ref{fig:moti:a}, this approach extracts transient components by introducing additional NeRF pipelines. Specifically, NeRF-W~\cite{wild21} and subsequent work NRW~\cite{sun2022neural} optimize appearance and transient embeddings through individual NeRF modules, rendering static fields and transient fields respectively.\textbf{(ii) Prior-Assisted NeRFs}. As illustrated in \figurename~\ref{fig:moti:b}, this approach leverages prior knowledge to assist in separating transient objects from the background. In particular, SF-NeRF~\cite{wild23semantic} introduced an occlusion filtering module to remove transient objects by a pretrained semantic segmentation model. While Ha-NeRF~\cite{wild21hal} eliminated transient components in pixel-wise by an anti-occlusion module image dependently. Despite showcasing promising results, these approaches still face problem that inaccurate transient decomposition from the complex scene and entanglement reconstruction of static appearances and occlusion.

In this paper, we address this problem from a fresh perspective by drawing inspiration from recent advances in image inpainting~\cite{image_inpaint1,image_inpaint2}, which aims to remove unwanted occluders and make imaginative restoration. The key insight of our work is that eliminating transient objects in NeRF is, in essence, an inpainting process during reconstruction. In addressing the mentioned challenges of NeRF, our objective is to perceive and separate undesired, blurry foreground areas in rendered images and produce plausible and consistent background scenes, which is exactly the expertise of image inpainting.


To this, we propose Inpainting Enhanced NeRF, or \textbf{IE-NeRF}, a novel approach that utilizes inpainting to separate transient content within NeRF. As illustrated in \figurename~\ref{fig:moti:c}, our model comprises three modules: the regular NeRF for static scene image rendering, the transient mask generator, and the inpainting module for removing transient components and repairing static images. Given an image, the model first encodes it into a high-dimensional vector using a CNN. Unlike the conventional NeRF, the MLP network in \ours\ predicts not only the color and voxel density, but also the masks for the transient components. The original image along with the masks are fed into the inpainting module to generate the restored static image. Finally, the rendered static image, obtained through volume rendering, is optimized by minimizing the photometric loss with the restored static image. Comprehensive experiments validate the performance of our method. 


Our contributions can be summarized as follows: 
\begin{itemize}
    \item We propose to enhance NeRF in uncontrolled environments by incorporating inpainting. Drawing on the success of the inpainting technique, this represents a novel approach compared to prior efforts.  
    
    \item We extend the MLPs network of NeRF to generate both static elements and transient masks for an image simultaneously. This allows us to take advantage of inpainting to restore the static scene image while eliminating occlusions with the transient prior, thereby contributing to the optimization of static rendering.
    
    \item To improve the learning of transient components, we introduce a training strategy that adopts frequency regularization with transient mask factor in integrated positional encoding. 
\end{itemize}

\section{Related Work}



\label{sec:relate}

\subsection{Novel View Synthesis}

Novel View Synthesis is a task that aims to generate new views of a scene from existing images~\cite{freeview}. NVS usually involves geometry-based image reprojection~\cite{unstructlum,deepblend} and volumetric scene representations~\cite{deepstereo,neural_volumes,chen2023neurbf}. The former applies techniques such as Structure-from-Motion~\cite{hartley2003multiple} and bundle adjustment~\cite{bundle_adjust} to construct a point cloud or triangle mesh to represent the scene from multiple images, while the latter focuses on unifying reconstruction and rendering in an end-to-end learning fashion. 

Inspired by the layered depth images, explicit scene representations such as multi-plane images~\cite{stereo_mag,singleview} and multiple sphere images~\cite{somsi,matryodshka} have also been explored. They use an alpha-compositing~\cite{composite} technique or learning compositing along rays to render novel views. In contrast, implicit representation learning techniques like NeRF~\cite{freeview,nerf,hu2023tri} exhibit remarkable capability of rendering novel views from limited sampled data. While NVS has made significant progress, challenges persist, especially in addressing occlusions and enhancing the efficiency of rendering complex scenes. 

\subsection{Neural Rendering}

Neural rendering techniques are now increasingly employed for synthesizing images and reconstructing geometry from real-world observations in scene reconstruction~\cite{nerf_sota,kerbl20233d}. Various approaches utilize image translation networks and different learning components, such as learned latent textures~\cite{defer_nerf}, meshes~\cite{free_form_rec}, deep voxel~\cite{deepvoxels}, 3D point clouds~\cite{pcd_multiplane_proj}, occupancy fields~\cite{implicit_surface}, and signed distance functions~\cite{deepsdf,zhuo2023graph,zhuo2022proximity}, to enhance realistic content re-rendering and reconstruction.

The prominent NeRF model utilizes a Multi-Layer Perceptron (MLP) to restore a radiance field. Subsequent research has aimed at extending NeRF's capabilities to dynamic scenes~\cite{scene_flow,wang2024mp,zhuo2022efficient}, achieving more efficient rendering~\cite{f2nerf,wang2024mp}, refining pose estimation~\cite{inerf}, and exploring few-shot view synthesis~\cite{singleview,li2024taming}. Noteworthy studies~\cite{wild21,wild21hal} addressed the problem of view synthesis using internet photo collections, which often include transient occlusions and varying illumination. Additional work~\cite{wild23semantic,li2024taming} focused on NeRF training in a few-shot setting. In contrast to these approaches, our work seeks to enhance view synthesis in the wild by separating dynamic and static scenes via image inpainting.

\begin{figure*}[htbp]
\centering
\includegraphics[width=1.0\linewidth]{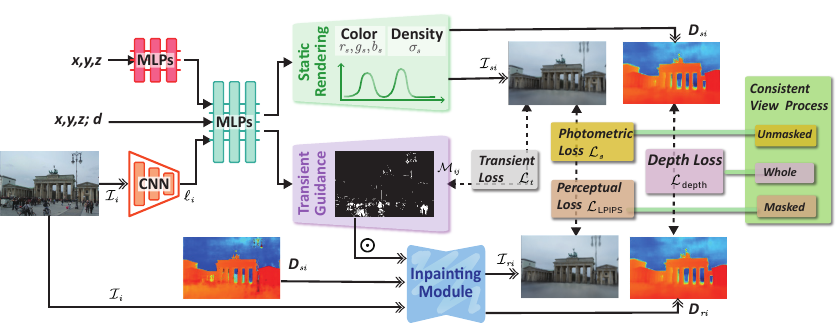}
\caption{
\ours\ framework: Given an image $\mathcal{I}_i$, a CNN is used to generate a feature embedding $\ell_i$. Then this embedding, along with the sample location $(x,y,z)$ and the view direction $d$ of the camera ray, is fed into MLPs to produce static color elements${r_s, g_s, b_s}$ and radiance intensity $\sigma_s$, as well as the transient mask embedding $\ell_t$, which is used to generate the transient mask $\mathcal{M}_{ij}$. The former is utilized to generate the new static scene image $\mathcal{I}_{si}$ through volumetric rendering, while the latter guides the Inpainting Module for the restoration of the static image $\mathcal{I}_{ri}$ and repaired depth map ${D}_{ri}$. Finally, we optimize the model while addressing view consistency by minimizing the photometric loss, the LPIPS loss, the depth loss, and the transient loss.
}
\label{fig:framework}
\end{figure*}

\subsection{Image Inpainting}
Image inpainting is a technique that fills in missing or damaged regions in an image, widely used for removing unwanted objects from images, and reconstructing deteriorated images~\cite{inpaint_survey,wang2024generating}. There have also been researches looking into integrating inpainting within NeRF~\cite{mirzaei2023spin,guo2023forward}. Liu~\textit{et al.}~\cite{nerfin} removed unwanted
objects or retouched undesired regions
in a 3D scene represented by a trained NeRF. 
Weder~\textit{et al.}~\cite{remove_obj} proposed a method that utilizes neural radiance fields for plausible removal of objects from output renderings, utilizing a confidence-based view selection scheme for multi-view consistency. 
Mirzaei~\textit{et al.}~\cite{refguide} utilized image inpainting to guide both the geometry and appearance, performing inpainting in an inherently 3D manner of NeRFs. In our research,
we investigate the synergistic effect of the inpainting module in joint learning within the NeRF pipeline. we use an inpainting module and the transient mask from NeRF to generate the inpainted static image, which helps guiding the optimization process for rendering the final reconstructed image.  

\section{Methodology}

\label{sec:meth}

\subsection{NeRF Preliminary}
\label{sec:pre}
NeRF models a continuous scene using a 5D vector-valued volumetric function $F(\theta)$ on $\mathbb{R}^3\times\mathbb{S}^2$, implemented as an MLP.
This function takes a 3D location $(x,y,z)\in\mathbb{R}^3$ and a 2D viewing direction $d=(\theta, \varphi)\in\mathbb{S}^2$ as input, produces an emitted color $(r,g,b)$ and volume density $(\sigma)$ as outputs. NeRF represents the volumetric density $\sigma(t)$ and color $c(t)$ at point of camera ray
$r(t)$ using MLPs with ReLU activation functions. Formally: 	
\begin{equation}
    \left[\sigma(t),z(t)\right]=\mbox{MLP}_{\theta_{1}}
    \left[\gamma_x(r(t))\right],
\end{equation}
\begin{equation}
    c(t)=\mbox{MLP}_{\theta_{2}}
    \left[\gamma_{d}(d),z(t)\right],
    \label{eq:ct}
\end{equation}
where $\theta=[\theta_1, \theta_2]$ represents the collection of learnable weights and biases of MLPs. The functions $\gamma_x$ and $\gamma_d$ are predefined encoding functions used for the spatial position and viewing direction, respectively. NeRF models the neural network using two distinct MLPs. The output of the second MLPs is conditioned on $z(t)$, one of the outputs from the first MLPs. This emphasizes the point that the volume density $\sigma(t)$ is not influenced by the viewing direction $d$. To calculate the color of a single pixel, NeRF uses numerical quadrature to approximate the volume rendering integral along the camera ray. The camera ray is represented as $r(t)=o+td$, where $o$ is the center of the projection and $d$ is the direction vector. NeRF's estimation of the expected color $\hat{C}_r$ for that pixel is obtained by evaluating the integral along the ray:
\begin{equation}
    \hat{C}(r)=\mathcal{R}(r,c,\sigma)=
        \sum\limits_{k=1}^{K}T(t_k)\cdot
        \alpha(\sigma (t_k)\delta_k)\cdot
        c(t_k),
    \label{eq:cr}
\end{equation}
\begin{equation}
    T(t_k)=\exp\left(
        -\sum\limits_{k'=1}^{k-1}{
            \sigma(t_{k'})\cdot\delta_{k'}
        }
    \right),
\end{equation}
where $\alpha(x)=1-\exp(-x)$, $(r,c,\sigma)$ signifies the volume radiance field along the ray $r(t)$, while $\sigma(t)$ indicates the density value and $c(t)$ represents the color at each point along the ray. $\delta_{k}=t_{k+1}-t_k$ denotes the distance between two integration points. For enhanced sampling efficiency, NeRF employs a dual MLP strategy: it comprises coarse and fine networks sharing the same architecture. The optimization of both models' parameters is achieved by minimizing the following loss function:
\begin{equation}
    \sum\limits_{ij}
    \left\|C(r_{ij})-\hat{C}_c(r_{ij})\right\|_2^2+
    \left\|C(r_{ij})-\hat{C}_f(r_{ij})\right\|_2^2,
    \label{eq:loss}
 \end{equation}
where $C(r_{ij})$ represents the observed color along the ray $j$ in the image $\mathcal{I}_i$, and $\hat{C}_c, \hat{C}_f$ are the predictions of the coarse and fine models, respectively.

\subsection{Pipeline Overview}
\label{sec:overview}

The pipeline of our model is illustrated in \figurename~\ref{fig:framework}. \chadded[id=R1]{To deal with the view-dependent appearance input, we propose to learn independent feature representations using a convolutional neural network-based encoder, denoted as \( E_\mathcal{E} \), which encodes each input image \( I_i \) into a corresponding feature latent vector $\ell_i$. This feature latent vector also captures the unique characteristics of each image's appearance, such as lighting conditions and scene variations.}  Then we use the core model of NeRF that consists of two MLP modules, namely $\mbox{MLP}_{\theta_1}$ and $\mbox{MLP}_{\theta_2}$, described as Eq.~\eqref{eq:MLP1} and Eq.~\eqref{eq:c_i(t)}.
\chdeleted[id =R1]{At the initial stage, the input image $I_i$ is processed through CNN producing a high-dimensional vector $\ell_i$.} 
The pose position $(x,y,z)$ of the image $I_i$ is processed through an $\mbox{MLP}_{\theta_1}$ with the output of $z_i(t)$. While the $\ell_i$ and  $z_i(t)$ combined with the original position $(x,y,z)$ and directional $d$ are fed into the second network $\mbox{MLP}_{\theta_2}$. 
\chadded[id=R1]{To incorporate these learned features into the radiance field model, we replace the radiance \( c(t) \) used in Eq.~\eqref{eq:c_i(t)} with an image-dependent radiance \( c_i(t) \) as introduced in Eq.~\eqref{eq:ct}. This modification introduces a dependency on the image index \( i \), effectively adapting the radiance to specific appearance features of each image. }

\begin{equation}
    \left[\sigma (t),z_i(t)\right]=
    \mbox{MLP}_{\theta_1}\left[\gamma_x(r(t))\right],
    \label{eq:MLP1}
\end{equation}
\begin{equation}	
    c_i(t)=\mbox{MLP}_{\theta_2}\left[\gamma_d(d),z_i(t)\right].
    \label{eq:c_i(t)}
\end{equation}

\chadded[id=R1]{Consequently, the approximated pixel color \( \hat{c}_i \) becomes dependent on \( i \), transitioning from the \( \hat{c} \) used in Eq.~\eqref{eq:cr} to the image-dependent approximation \( \hat{c}_i \) described in Eq.~\eqref{eq:c_i(r)}.}
$\mbox{MLP}_{\theta_2}$ outputs two components. The first part includes the static color $(r_s, g_s, b_s)$ and radiance intensity $\sigma_{s}$. These elements are used to generate a new static scene image $I_s$ through volumetric rendering. We construct the volumetric rendering formula based on Eq.~\eqref{eq:cr},

\begin{equation}
    \hat{C}_i(r)=\mathcal{R}(r,c,\sigma )=
    \sum\limits_{k=1}^{K}{
        T(t_k)\cdot\alpha(\sigma (t_k)\delta_k)\cdot c_i(t_k)
    },
    \label{eq:c_i(r)}
\end{equation}
where $\sigma(t)$ is the static density and $c_{i}(t)$ is the color radiance. The second part of the output of $\mbox{MLP}_{\theta_2}$ is the transient mask embedding $\ell_i$, which is used to guide the inpainting module for restoring the static scene during training. After the original image $I_i$ is inpainted with the transient mask $M_{ij}$, we obtain the restored static image $\mathcal{I}_{ri}$ and depth map ${D}_{ri}$, which acts as supervision for the optimization of the rendered static image $\mathcal{I}_{si}$ and depth map ${D}_{si}$. 

\subsection{Transient Masks and Inpainting Module}
\label{sec:transient}

We take advantage of the MLPs to generate the transient mask representation and the pre-trained inpainting model for the inpainting task. The mask is designed to capture dynamic elements in a scene, such as moving objects or changing conditions. Rather than relying on a 3D transient field to reconstruct transient elements specific to each image, as adopted in NeRF-W~\cite{wild19}, we use an image-dependent 2D mask map to eliminate the transient effects.
\chdeleted[id=R1]{The transient mask is generated along with the regular output of density and color radiance from the same MLPs as $\mathcal{M}_{ij}$. 
The mask maps a 2D-pixel location $l=(u,v)$ and an image-specific transient image embedding $\ell_i$ to a 2D pixel-wise probability representation as follows:}
\chadded[id=R1]{The image-independenit transient mask embedding is generated along with the regular output of density and color radiance from the same MLPs as $\ell_t$, the embedding then is processed by additional linear layers $E_\phi$, which maps a 2D-pixel location $l=(u,v)$ and the transient image embedding $\ell_t$ to a 2D pixel-wise probability representation as $\mathcal{M}_{ij}$, the equation is as follows:}
\begin{equation}	
\mathcal{M}_{ij}=E_\phi(l_{ij}, \ell_t).
\end{equation}  

Once the transient mask is generated, it serves as guidance for the inpainting module. Here we use the inpainting model LaMa~\cite{robust_inpaint}, which is the state-of-the-art single-stage image inpainting system and is robust to large masks with less trainable parameters and inference time. LaMa performs inpainting on a color image $x$ that has been masked by a binary mask $m$, denoted as $x\odot m$. The input to LaMa is a four-channel tensor $x'=stack(x\odot m,m)$, where the mask $m$ is stacked with the masked image $x\odot m$. Taking $x'$, the inpainting network processes the input using a feed-forward network $G_\Delta$ in a fully convolutional manner and produces an inpainted image $\hat{x}=G_{\Delta}(x')$. In our pipeline, the original input image $I_{i}$, along with the thresholded transient mask $m_{ij}$, is fed into the pre-trained inpainting module. With the guidance of $M_{ij}$, LaMa accurately separates the transient components from the original image, then restores and repairs the features of the static scene. This process results in the repaired static scene image $I_{ri}$, which acts as the supervision for the static rendered image.  

\subsection{Loss Function and Optimization}
\label{sec:loss}
\chadded[id=R1]{
Directly using the inpainting result from off-the-shelf inpainting module as supervision may cause blurry results, since the inpainting module’s capability of recovering accurate raw images is not yet fully perfected, even with the state-of-the-art method as demonstrated in \cite{robust_inpaint}. In addition, the potential inconsistencies across different views also have a negative impact on reconstruction\cite{weder2023removing}.
To capitalize on strengths and mitigate weaknesses of inpainting, we designed the static photometric loss to optimize the rendered static scene image $I_{si}$ from two aspects: one is supervised by the ground truth colors which are outside the transient mask using mean squared error, and the other is supervised inside the transient mask with a perceptual loss, LPIPS\cite{zhang2018unreasonable}.
The photometric mean squared error is as follows:
\begin{equation}
    \mathcal{L}_s=
    \left\|
        {C}_s(r_{ij}) - \hat{C}_{ci}(r_{ij}) 
    \right\|_2^2 + 
    \left\| 
        {C}_s(r_{ij}) - \hat{C}_{fi}(r_{ij}) 
    \right\|_2^2,
    \label{eq:Ls}
\end{equation}
where ${C}_s(r_{ij})$ represents the true color of ray $j$ for the rendered static scene image $I_s$ in image $I_i$. $\hat{C}_{ci}(r_{ij})$ and $\hat{C}_{fi}(r_{ij})$ represent the color estimate derived from the coarse and the fine model, respectively.
The perceptual loss, LPIPS is as follows:}

\begin{equation}
    \label{eq:perceptual.loss}
        \mathcal{L}_\text{LPIPS} = \sum_{i \in \mathcal{B}} \text{LPIPS}( I_{si}, I_{ri}),
\end{equation}
where $\mathcal{B}$ is the unmasked area of the $i$-th view image, and $I_{si}$ is the rendered from NeRF, $I_{ri}$ is generated by inpainting module.


\chadded[id=R1]{Except for the color optimization, we also consider the depth for view consistency.  Similar as\cite{mirzaei2023spin} done.
we use the static rendering pipeline to generate the depth images $\{D_{si}\}_{i=1}^{n}$, corresponding to the training views. Depth maps are created by using the distance of points to the camera in place of the color, similar as Eq.~\ref{eq:c_i(r)}: }

\begin{equation}
    \hat{D}_{si}(r)=\mathcal{R}(r,c,\sigma )=
    \sum\limits_{k=1}^{K}{
        T(t_k)\cdot\alpha(\sigma (t_k)\delta_k)\cdot c_i(t_k)
    },
    \label{eq:D_i(r)}
\end{equation}

\chadded[id=R1]{The rendered static depths are then given to the inpainting module to obtain restored depth maps $\{\hat D_{ri} \}_{i=1}^{n}$ under the guidance of the transient mask. The inpainted depth maps are then used to supervise the static NeRF's geometry, via the $\ell_2$ distance of its rendered depths $\hat D_{si}$, to the inpainted depths $D_{ri}$:}
\begin{equation}
    \label{eq:depth.loss}
    \mathcal{L}_\text{depth} = \frac{1}{\vert \mathcal{R}\vert} \sum_{r \in \mathcal{R}} 
    \left| \hat D_{si}(r) - \ D_{ri}(r) \right|^2,
\end{equation}

where $\hat D_{si}(r)$ and $D_{ri}(r)$ are the depth values for a ray $r$, $\mathcal R$ is a ray batch sampled from the training views.

In addition to the static loss, we also consider the transient components. 
\chadded[id=R2]{As described in Section 3.3, we gain the transient mask as 2D
pixel-wise probability representation $\mathcal{M}_{ij}$ from NeRF MLPs $F_\theta$ and additional linear layers network $E_\phi$.}
The transient image is derived from the original image using the transient mask probability map indicating the visibility of rays originating from the static scene. To separate static and transient components, we optimize the mask map during the training process in an unsupervised manner. Thus, we provide the transient loss as follows: 

\begin{align}
\mathcal{L}_t = &(1 - \mathcal{M}_{ij}) \left\| {C}(r_{ij}) - \hat{C}_s(r_{ij}) \right\|_2^2 \nonumber \\
&+ \lambda \mathcal{M}_{ij} \left\| {C}_s(r_{ij}) - \hat{C}_s(r_{ij}) \right\|_2^2.
\label{eq:Lt}
\end{align}

Specifically, the first term tackles the occlusion error by taking into account transient when comparing the rendered static image with the original image.\chadded[id=R3]{ 
For rendered static scene color $\hat{C}_s(r_{ij})$,
we use the ground truth color ${C}_s(r_{ij})$ outside the transient mask and the repaired pixels' LPIPS inside the mask as supervisions, while here we rely on the original input color and the rendered color to deal with the existence of transient.}
The second term in Eq.~\eqref{eq:Lt} addresses the reconstruction error of static components between the rendered and repaired ground truth colors, under the assumption that the value of $\mathcal{M}_{ij}$ belongs to the static phenomena. The parameter $\lambda$ is used to adjust the balance between the transient and static components, helping to avoid the neglect of either phenomenon.
Then, we can obtain the final optimizing function by combining the loss terms with weight $\alpha$, $\beta$, $\gamma$ and $\rho$ respectively:
\begin{equation}
    \mathcal{L}=
    \alpha\sum\limits_{ij}{\mathcal{L}_s}
    +\beta\sum\limits_{ij}{\mathcal{L}_t}
+\gamma\sum\limits_{i}\mathcal{L}_\text{LPIPS}
+\rho\sum\limits_{i}\mathcal{L}_\text{depth}.
\end{equation}




\begin{figure}[t]
\centering
\subfloat[Visualizations of the positional encoding (PE) and integrated positional encoding (IPE).]
{
    \includegraphics[width=0.9\linewidth]{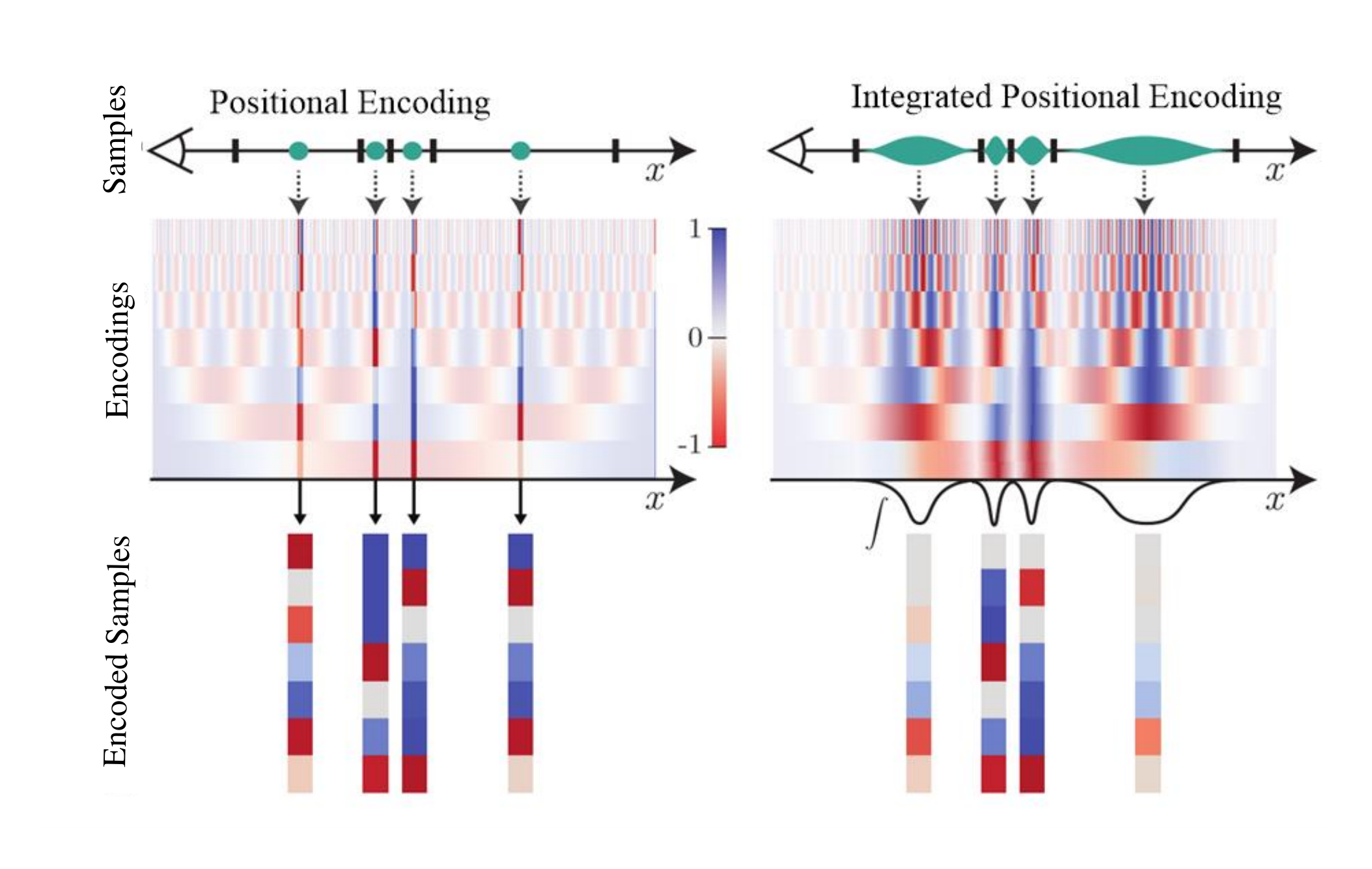}
    \label{fig: integrated positional encoding}
}
\\
\subfloat[Frequency regularization over training steps in one epoch.]{
    \includegraphics[width=0.8\linewidth]{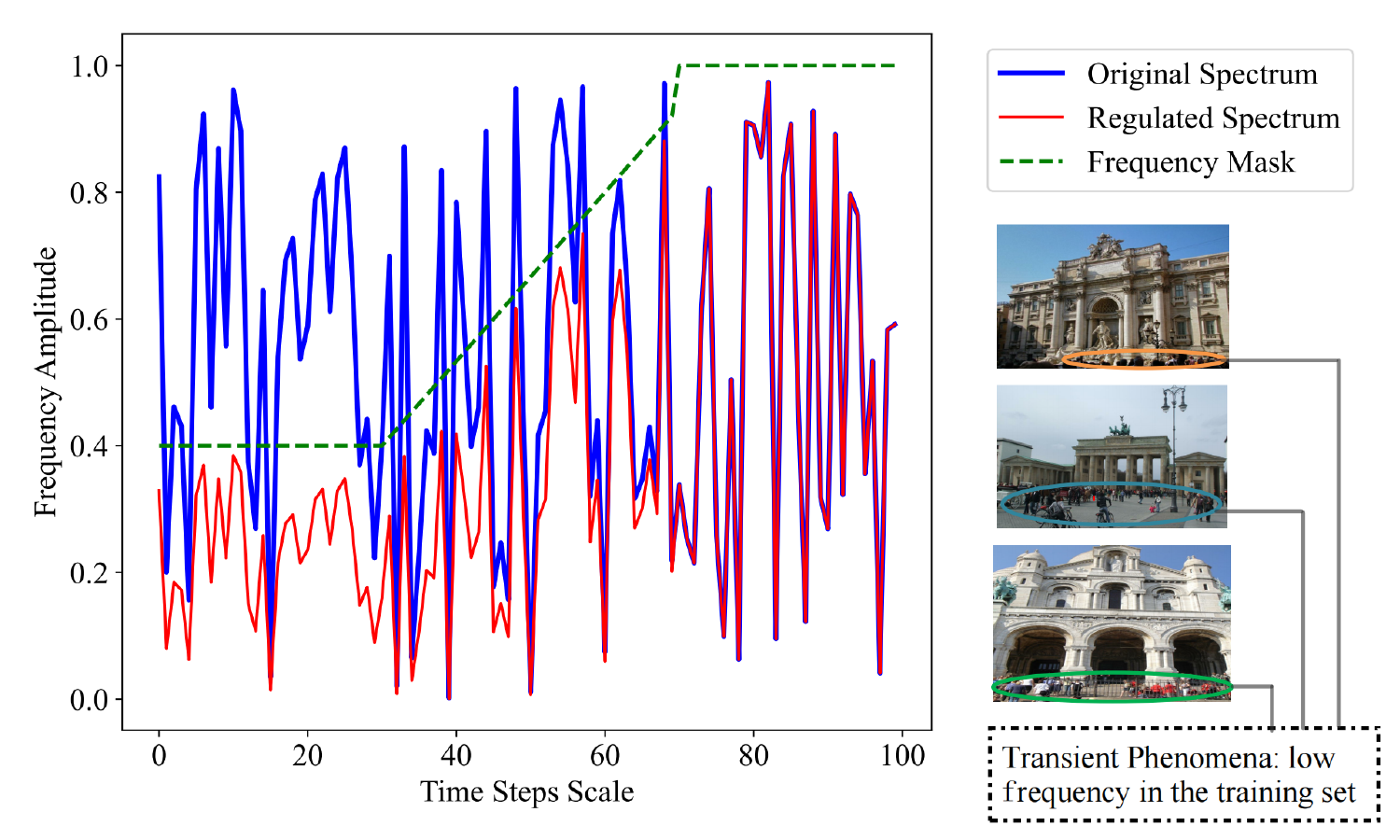}
    \label{fig:freq_reg}
}
\caption{
   Training strategy: Frequency regularization with integrated positional encoding. (a) Visualizations of the positional encoding methods. (b) We use a step piece-wise linearly increasing frequency mask (marked as the dotted green line) to regulate the frequency spectrum based on the training time steps.
}
\label{fig:training_strategy}
\end{figure}

\subsection{Training Strategy with Frequency Regularization}
\label{sec:train} 

To optimize the training process, we design a frequency regularization scheme with position integral encoding. Instead of employing a single ray per pixel, as done in NeRF, our approach utilizes mip-NeRF~\cite{mipnerf}, casting a cone whose radius adapts to variations in image resolution. This alteration transforms the positional encoding scheme from encoding an infinitesimally small point to integrating within the conical frustum (Integrated Positional Encoding) for each segment of the ray. This encoding method not only enables NeRF to learn multiscale representations but also demonstrates a performance where the participation of high-frequency signals in the encoding gradually increases as the training progresses~\cite{freenerf,barron2022mip}.

For unrestricted scenarios, the presence of transient elements often indicates a low density of low-frequency signals in the training set. It is noteworthy that the frequency of input can be regulated by position encoding. To facilitate the transient components' separation during the initial stages of training, we design a training strategy that initiates the process with raw inputs devoid of positional encoding and incrementally boosts the frequency amplitude in each iteration 
as training progresses. \chadded[id=R2]{In detail, We define the transient mask factor $\boldsymbol{r}$ as the ratio of transient pixels to the total image. To regulate the visible frequency spectrum during training, we use a linearly increasing frequency mask $\boldsymbol{\omega}$, which is applied under $\boldsymbol{r}$ and adjusts according to the training time steps as follows:}

\begin{gather}
    \gamma_L'(t,T;\mathbf{x}) = \gamma_L(\mathbf{x})\odot\boldsymbol{\omega}(t, T, L), \\
    \resizebox{0.8\linewidth}{!}{$ \displaystyle
        \mathrm{with}~~\boldsymbol{\omega}_i(t,T,L) = 
        \begin{cases}
            \displaystyle \boldsymbol{r} & \text{if}~~ i \leq \frac{t\cdot L}{T} + 3  \\
            \displaystyle (\frac{t\cdot L}{T} - \lfloor\frac{t\cdot L}{T}\rfloor)\cdot\boldsymbol{r} & \text{if}~~ \frac{t\cdot L}{T} + 3 < i \leq \frac{t\cdot L}{T} + 6  \\
            0 & \text{if}~~ i > \frac{t\cdot L}{T} + 6
        \end{cases}
    $} \label{eq:mask}
\end{gather}

\chadded[id=R2]{where $\gamma_L(\mathbf{x})$ is the positional encoding function, $L$ is the hyperparameter that determines the maximum encoded frequency, $\boldsymbol{\omega}_i(t, T, L)$ denotes the $i$-th bit value of $\boldsymbol{\omega}(t, T, L)$;  The variables $t$ and $T$ represent the current training iteration and the final iteration of frequency regularization, respectively.}

This frequency regularization strategy mitigates the instability and vulnerability associated with high-frequency signals, resulting in the separation of transient components during the initial stages of training. Furthermore, the early transient separation mask facilitates adjustments in the subsequent stages of training progress, gradually enhancing NeRF with high-frequency information and preventing both over-smoothing and interference with transient phenomena in the static scene reconstruction. The training strategy is illustrated in \figurename~\ref{fig:training_strategy}.



 \section{Experimental Results}
\subsection{Experimental Settings}

\label{sec:exp}
\paragraph{Datasets:} Our approach is evaluated on unconstrained internet photo collections highlighting cultural landmarks from the Phototourism dataset. We reconstruct four training datasets based on scenes from Brandenburg Gate, Sacre Coeur, Trevi Fountain, and Taj Mahal. 
\chadded[id=R2]{The dataset includes challenging in-the-wild photo collections of cultural landmarks, characterized by distinct environmental features such as blue skies and sunshine at the “Brandenburg Gate,” lush greenery at “Sacre Coeur,” light reflections at the “Trevi Fountain,” and both greenery and water reflections at the “Taj Mahal.” These scenes cover a wide range of lighting conditions, from sunny to overcast, and from daytime to nighttime, ensuring that our model can operate robustly across varying lighting scenarios. As an initial pre-filtering step, we discard low-quality images, specifically those with a NIMA\cite{tancik2020fourier} score below 3, to focus on clearer and more relevant data. Additionally, we filter out images where transient objects occupy more than 80 \% of the frame, following the methodology used in NeRF-W \cite{wild21}, utilizing a DeepLab v3\cite{chen2017rethinking} model trained on the Ade20k dataset.
}
We perform the train set and test set split using the same approach as employed by HA-NeRF~\cite{wild21hal}. Additionally, we downsample the original images by a ratio of 2 during training, consistent with the approach taken by NeRF-W and HA-NeRF.

\paragraph{Implementation Details:} 
Our implementation of the NeRF in the wild network is structured as follows: The entire neural radiance field consists of eight fully connected layers with 256 channels each, followed by two different activation tasks, one is ReLU activations to generate $\sigma$, the other is sigmoid activation following another 128 channels connected layer, to generate the transient mask possibility $M_{ij}$. Additionally, there is one more fully connected layer with 128 channels and a sigmoid activation, responsible for outputting the static RGB color $c$. For the image inpainting module, we utilize the LaMa model~\cite{robust_inpaint}, a pre-trained repairer employing a ResNet-like architecture with 3 downsampling blocks, 12 residual blocks using Fast Fourier Convolution (FFC), and 3 upsampling blocks. 

\paragraph{Baselines:} We evaluate our proposed method against several state-of-the-art NeRF models in the wild, including NeRF\cite{nerf}, NeRF-W\cite{wild21} and HA-NeRF\cite{wild21hal}. \chadded[id=R2]{Specifically, NeRF\cite{nerf} synthesizes novel views from 2D images by learning a volumetric scene representation, excelling in static scenes with fixed lighting but struggling in dynamic environments or under changing illumination due to its reliance on a static radiance field. NeRF-W\cite{wild21} extends NeRF to reconstruct realistic scenes from tourism images with varying appearances and occlusions. It achieves this by learning a per-image latent embedding that captures photometric appearance variations and utilizing a 3D transient field to model transient objects. Compared to NeRF-W, HA-NeRF\cite{wild21hal} consistently hallucinates novel views with unlearned appearances, effectively addressing time-varying appearances and mitigating transient phenomena using an image-dependent 2D visibility map}. 
To ensure a fair comparison, we maintain consistency in the main NeRF architecture across all models. This architecture comprises 8 layers with 256 hidden units for generating density $\sigma$, along with an additional layer of 128 hidden units for color $c$.

%
\begin{figure*}[htbp]
\centering
\includegraphics[width=\linewidth]{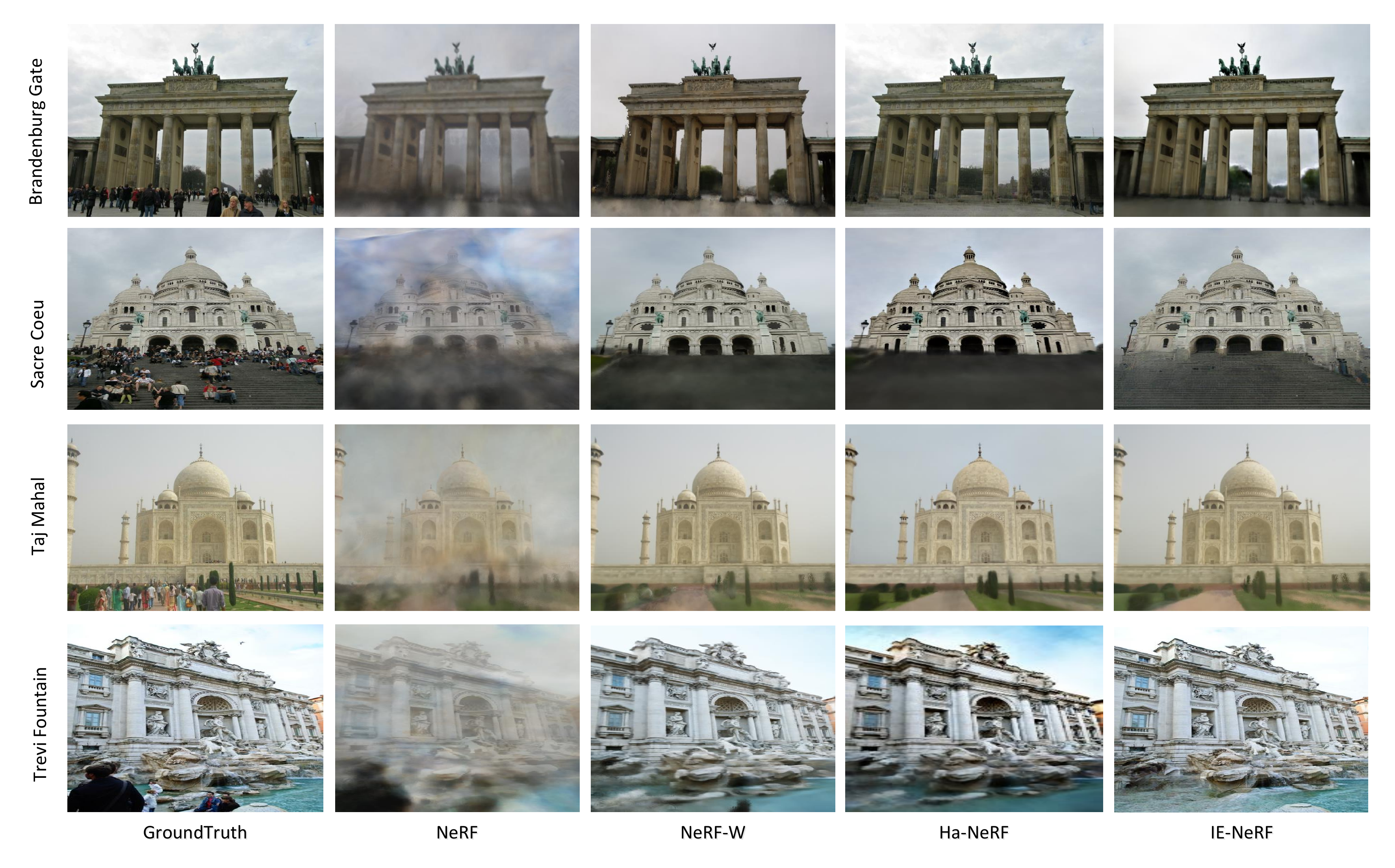}
\caption{Visual comparison results on four scenes of phototourism constructed datasets. \ours\ can remove the transient occlusions more naturally and render a consistent 3D scene geometry with finer details than existing methods.}
\label{fig:viusal_compare}
\end{figure*}
\begin{table*}[!t]
\caption{Comparison of PSNR, SSIM, and LPIPS for \ours\ and other models: NeRF, NeRF-W, Ha-NeRF on Phototourism datasets across specific scenes (Brandenburg Gate, Sacre Coeur, Trevi Fountain, and Taj Mahal).}
\centering
\resizebox{\linewidth}{!}{  
\begin{tabular}{lrrrrrrrrrrrr}
\toprule
    &\multicolumn{3}{c}{Brandenburg Gate} & \multicolumn{3}{c}{Sacre Coeur} &  \multicolumn{3}{c}{Trevi Fountain} & \multicolumn{3}{c}{Taj Mahal} \\
\cmidrule{2-13}
    & PSNR$\uparrow$ & SSIM$\uparrow$ & LPIPS$\downarrow$  
    & PSNR$\uparrow$ & SSIM$\uparrow$ & LPIPS$\downarrow$  
    & PSNR$\uparrow$ & SSIM$\uparrow$ & LPIPS$\downarrow$  
    & PSNR$\uparrow$ & SSIM$\uparrow$ & LPIPS$\downarrow$  \\
\midrule
    NeRF & 18.90 & 0.816 & 0.232 & 15.60 & 0.716 & 0.292 & 16.14 &0.601 & 0.366 &15.77 &0.697 &0.427\\
    NeRF-W & 24.17 &0.891 & 0.167 & 19.20 & 0.807 & 0.192 & 18.97 & 0.698 & 0.265  &$\textbf{26.36}$ &$\textbf{0.904}$ &0.207\\
    Ha-NeRF & 24.04 & 0.877 & \textbf{0.139} & 20.02 & 0.801 & 0.171 & 20.18 &0.691 &0.223 & 19.82 &0.829 &0.243\\
    SF-NeRF (30-fews) &23.23 &0.846 &0.178 &19.64 &0.757 &0.186 &20.24 &0.657 &0.243 &20.86 &0.820 &0.218 \\
\midrule
    \ours\ (Ours) &\textbf{25.33} & \textbf{0.898} & 0.158 &$\textbf {20.37}$ & \textbf{0.861} & $\textbf{0.169}$ & $\textbf {20.76}$ & $\textbf{0.719}$ & $\textbf{0.217}$ & 25.86 & 0.889 & $\textbf{0.196}$ \\
\bottomrule
\end{tabular}}
\label{tab:comparison}
\end{table*}
\begin{figure*}[htbp]
    \centering
    \includegraphics[width=0.85\linewidth]{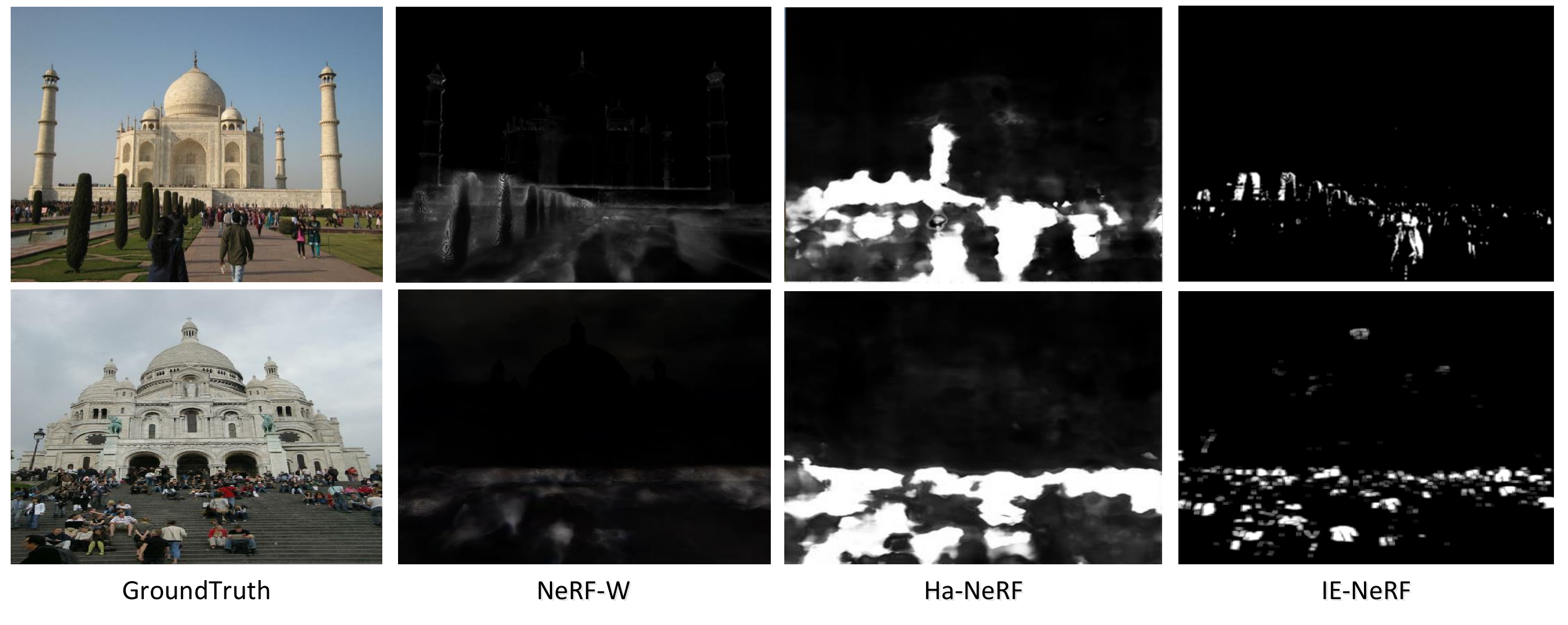}
    \caption{Comparisons of transient components predicted by \ours\ and baselines. The predicted transient components of NeRF-W are rendered with the 3D transient field, while Ha-NeRF predicts the transient visibility map with the pre-trained MLP.}
    \label{fig:transient_mask}
\end{figure*}

\paragraph{Evaluation:} The performance of our \ours\ and baselines are assessed by utilizing a held-out image and its associated camera parameters. After rendering an image from the matching pose, we evaluate its similarity with the ground truth. We provide a set of standard image quality metrics to assess the performance of the models. These metrics include Peak Signal Noise Ratio (PSNR), measuring the fidelity of the reconstructed image; Structural Similarity Index Measure (SSIM), evaluating the structural similarity between the generated and ground truth images; and Learned Perceptual Image Patch Similarity (LPIPS), which leverages perceptual image similarity through insights derived from learned features.

\subsection{Results Comparision}
\paragraph{Quantitative Results:}
We conduct experiments to compare the performance of our proposed method with existing baselines on the Phototourism dataset, specifically focusing on scenes of Brandenburg Gate, Sacre Coeur, Trevi Fountain, and Taj Mahal. The four scenes present unique challenges and variations in lighting conditions with transient phenomena. The quantitative results are summarized in \tablename~\ref{tab:comparison}, where we report PSNR/SSIM (higher is better) and LPIPS (lower is better). As depicted in \tablename~\ref{tab:comparison}, our proposed model, \ours, mostly demonstrates superior performance compared to the baselines. The reported PSNR and SSIM values, indicative of image fidelity and structural similarity, show that our model achieves higher scores, highlighting its effectiveness in reconstructing static image quality. Additionally, the lower LPIPS scores further emphasize the superiority of \ours\ in minimizing perceptual differences compared to the baselines.

\begin{figure*}[htbp]
    \centering    \includegraphics[width=\linewidth]{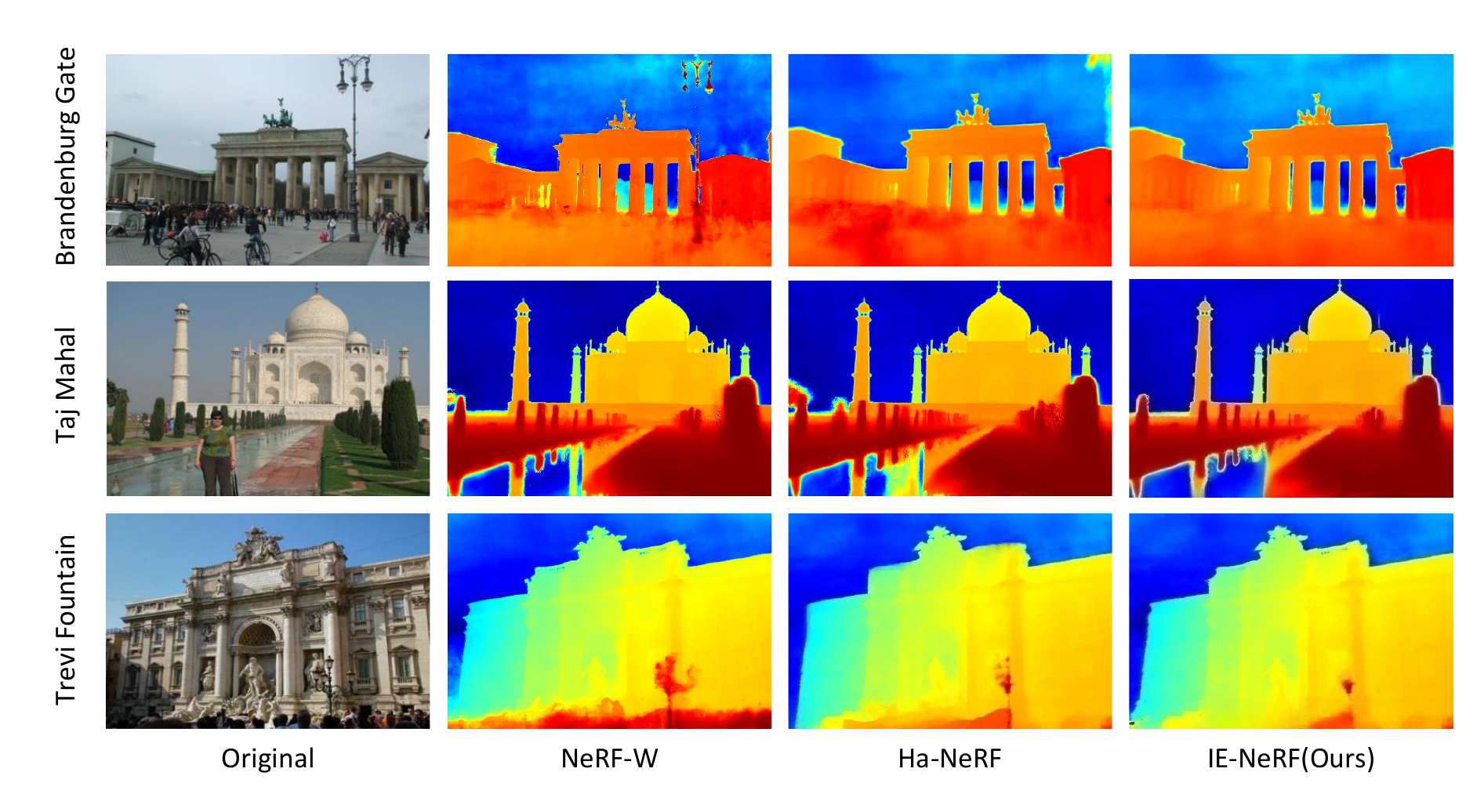}
    \caption{Samples of predicted static depth images of our \ours\ and other baselines on the scenes of the Phototourism datasets.}
    \label{fig:depth images}
\end{figure*}

\paragraph{Qualitative Results:}
We obtain qualitative comparison results based on different scenes of our model and the baselines. The rendered static images are shown in \figurename~\ref{fig:viusal_compare}, showing that the rendering process with NeRF is challenged by the persistence of transient phenomena, leading to global color deviations and shadowing effects. While NeRF-W and HA-NeRF demonstrate the capability to model diverse photometric effects, facilitated by the incorporation of appearance embeddings, it is important to note that they can not avoid the rendering of ghosting artifacts in Tajcompari Mahal and Brandenburg Gate (seriously on NeRF-W) and blurry artifacts in Trevi Fountain and Sacre Coeur. On the contrary, \ours\ disentangles transient elements from the static scene consistently which proves the effectiveness of the transient mask-guided inpainting module. In addition, we present the transient components of NeRF-W, Ha-NeRF, and \ours\ in \figurename~\ref{fig:transient_mask}. Results from \ours\ reveal more detailed texture in the pixel mask. Transient performance further supports the capability of our \ours\ method.
\begin{figure*}[htbp]
    \centering
    \includegraphics[width=\linewidth]{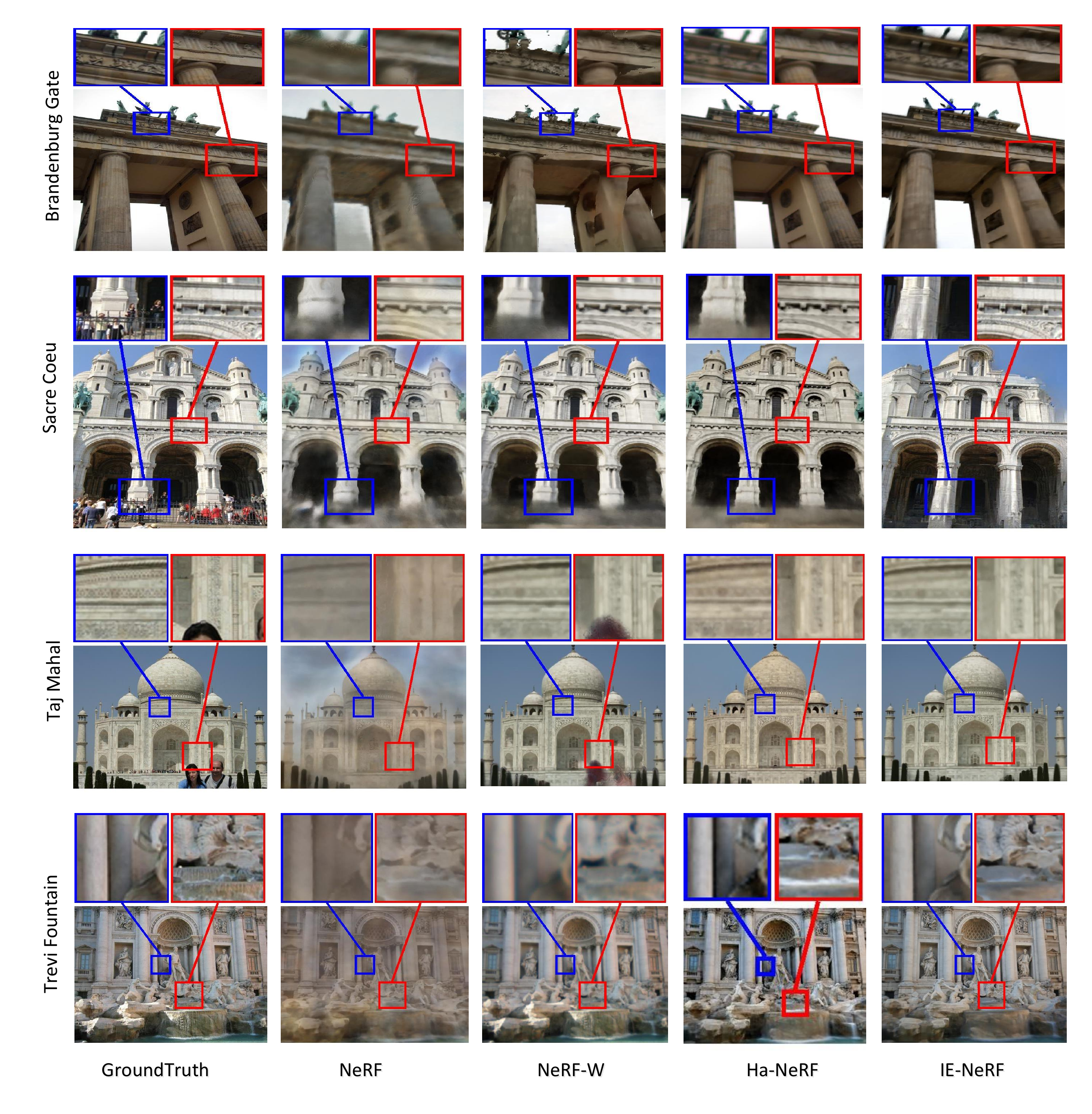}
    \caption{Further qualitative results from experiments on the Phototourism dataset reveal the effectiveness of IE-NeRF in addressing transient phenomena under diverse conditions. This includes handling various reflections in the top and bottom rows, maintaining consistent scene geometry at a far distance, and proficiently eliminating transient occluders in the middle rows.}
    \label{fig:Further qualitative results under diverse conditions }
\end{figure*}
 Qualitative results and zoomed-in views of rendered static image from experiments on the Phototourism dataset are presented in \figurename~\ref{fig:Further qualitative results under diverse conditions }. Our method not only performs well in removing transient phenomena but also excels in reconstructing static images with more details and high fidelity. Furthermore, as observed in static depth images \chdeleted[id=R1]{and rendered static color images} across scenes in \figurename~\ref{fig:depth images}, our method provides less information about transient occluders and clearly details compared to NeRF-W and Ha-NeRF, particularly in reflection of water surface in Taj Mahal scene. These findings further support the capability of \ours.

\subsection{Ablation Studies}
In this section, we conduct ablation studies exploring diverse transient mask generation approaches with the same pipeline of static image reconstruction. We compare three methods: (1) \ours(ours), our primary approach utilizing a transient mask generated by the NeRF MLP network. (2) \ours(IM), which generates the transient mask with an independent MLP network, following a way similar to the Ha-NeRF~\cite{wild21hal}. (3) \ours(SM), where we leverage a pre-trained semantic model, specifically the object instance segmentation model MaskDINO~\cite{r52}, in which we predefine the transient objects, including but not limited to people, cars, bicycles, flags, and slogans. We evaluate the ablation models on Phototourism datasets of PSNR, SSIM, and LPIPS metrics and provide the results in \tablename~\ref{tab:ours_ablation}. 

\begin{table}[!t]
\centering
\caption{Ablation comparison on metrics of PSNR, SSIM, and LPIPS results of different transient mask generation methods.}
\begin{tabular}{lrrr}
    \toprule
    Method & PSNR$\uparrow$ & SSIM$\uparrow$ & LPIPS$\downarrow$ \\
    \midrule
    \ours(IM) & 22.27 & 0.802 & 0.191 \\
    \ours(SM) & 24.65 & 0.879 & 0.187\\
    \ours(Ours) &\textbf{25.33} & \textbf{0.898} & \textbf{0.158}  \\
    \bottomrule
\end{tabular}
\label{tab:ours_ablation}
\end{table}
\begin{figure}[!t]
\centering
\includegraphics[width=\linewidth]{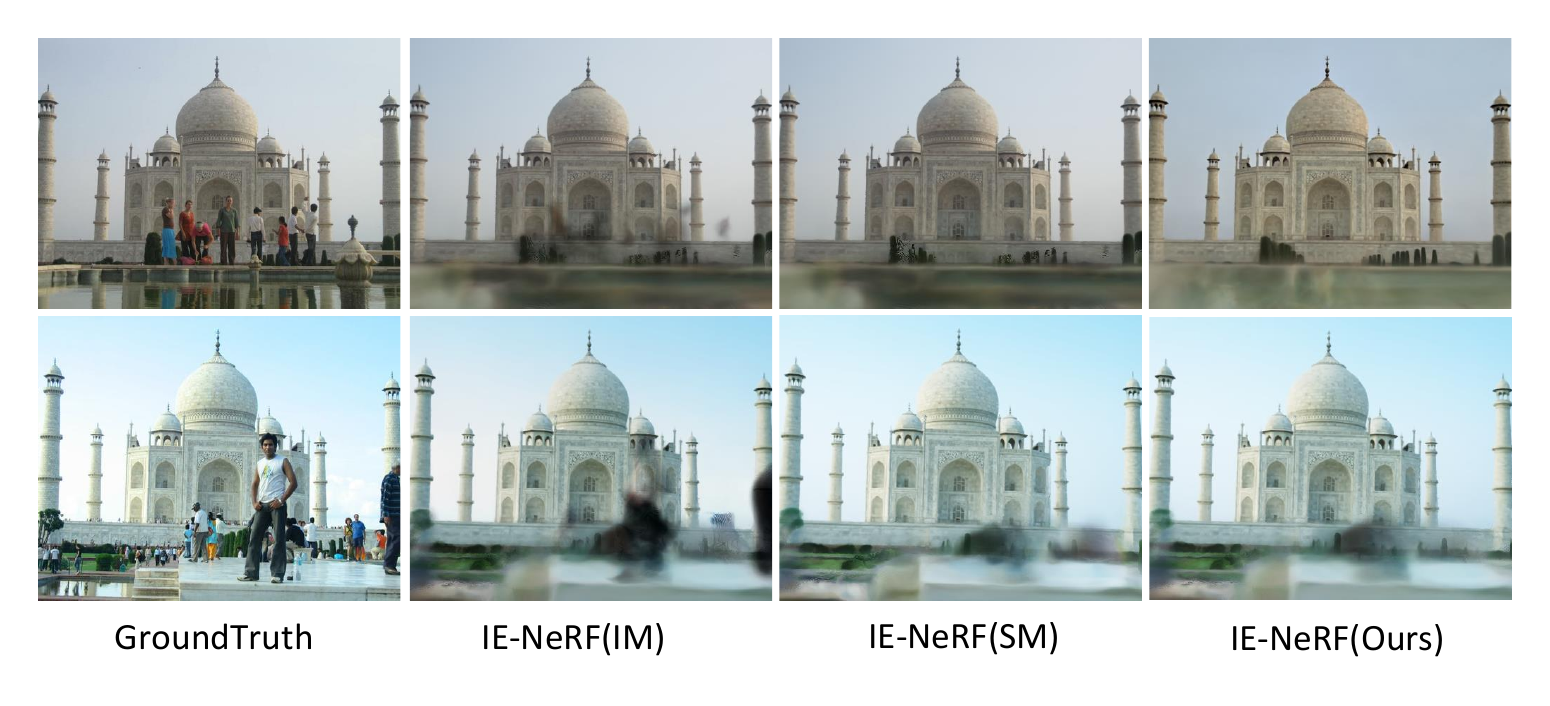}
\caption{Qualitative results of the ablation studies.}
\label{fig:ablation study}
\end{figure}

\chadded[id=R1]{The qualitative results are also shown \figurename~\ref{fig:ablation study}, \ours(IM) is less sensitive to occluders that are similar in color to the scene and tend to produce artifacts. Due to the absence of prior supervision and the inherent randomness of transient phenomena, the independent network, lacking spatial constraints, encounters difficulties in accurately delineating occlusion masks. Furthermore, \ours(SM) demonstrates limited effectiveness in removing smaller occluders that possess partial semantic features. This challenge arises from its reliance on predefined classes, which may overlook or inadequately address exceptional objects such as shadows or images on the wall.
IE-NeRF utilizes the transient mask generated by NeRF MLPs, which is further refined through additional neural layers supervised by the rendering pipeline. This approach leverages inpainting and rendering techniques to achieve optimization of the transient masks in both image and depth spaces, resulting in better novel view synthesis and static scene reconstruction. }  


\begin{figure}[!t ]
\centering
\includegraphics[width=0.75\linewidth]{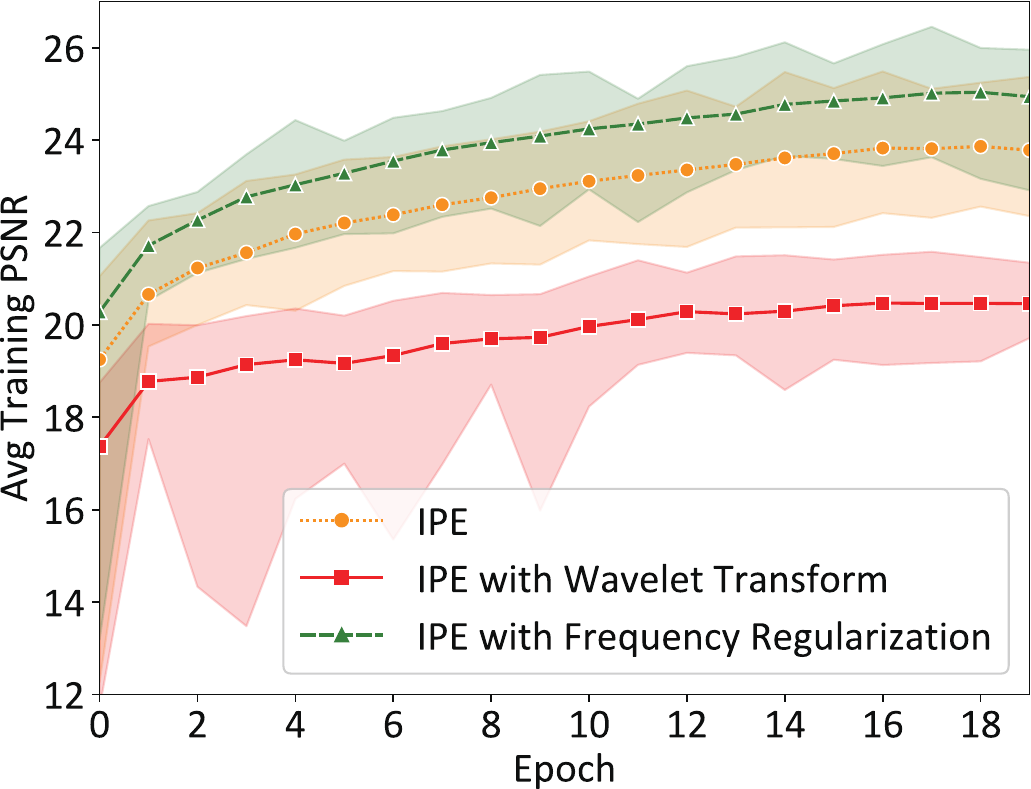}
\caption{
Comparisons of Training PSNR across epochs were conducted with different training strategies, including Integrated Positional Encoding (IPE), Integrated Positional Encoding with wavelet transform (WT-IPE), and Integrated Positional Encoding with frequency regularization (RegFre-IPE).
}
\label{fig:psnr_epoch}
\end{figure}

\begin{table}[!t]
\caption{Ablation comparisons of PSNR, SSIM, and LPIPS for IE-NeRF with different training strategies on Phototourism datasets in the scene of Brandenburg Gate.}
\centering
\begin{tabular}{lrrr}
\toprule
Training strategy & PSNR$\uparrow$ & SSIM$\uparrow$ & LPIPS$\downarrow$\\
\midrule
IPE & 23.58 & 0.864 & 0.191 \\
WT-IPE & 19.86  & 0.723 & 0.289 \\
RegFre-IPE & \textbf{25.33} & \textbf{0.898} & \textbf{0.158} \\
\bottomrule
\end{tabular}
\label{tab:train_ablation}
\end{table}

We also conduct ablation studies to analyze the effectiveness of our training strategy. We evaluate the performance on the scene of Brandenburg Gate datasets in the following ways: First, we employed the conventional approach using integrated positional encoding. Second, we implemented our proposed strategy, which incorporates regularization frequency into integrated positional encoding (RegFre-IPE). Third, we experimented with the wavelet transformer during the embedding process, considering the non-stationarity of input signals caused by transient phenomena. Comparisons of training PSNR across epochs with the three training approaches are shown in \figurename~\ref{fig:psnr_epoch}. The results indicate that RegFre-IPE achieves the highest PSNR in the early epochs and maintains a lead throughout the entire training period. However, integrated positional encoding with wavelet transform (WT-IPE) has negative impacts on performance. The metric results are summarized in \tablename~\ref{tab:train_ablation}. Our proposed training strategy with RegFre-IPE consistently demonstrates the best performance across all three considered metrics under the same training epochs.



\subsection{Other Discussions }

\begin{figure*}[htbp]
\centering
\includegraphics[width=0.9\linewidth]{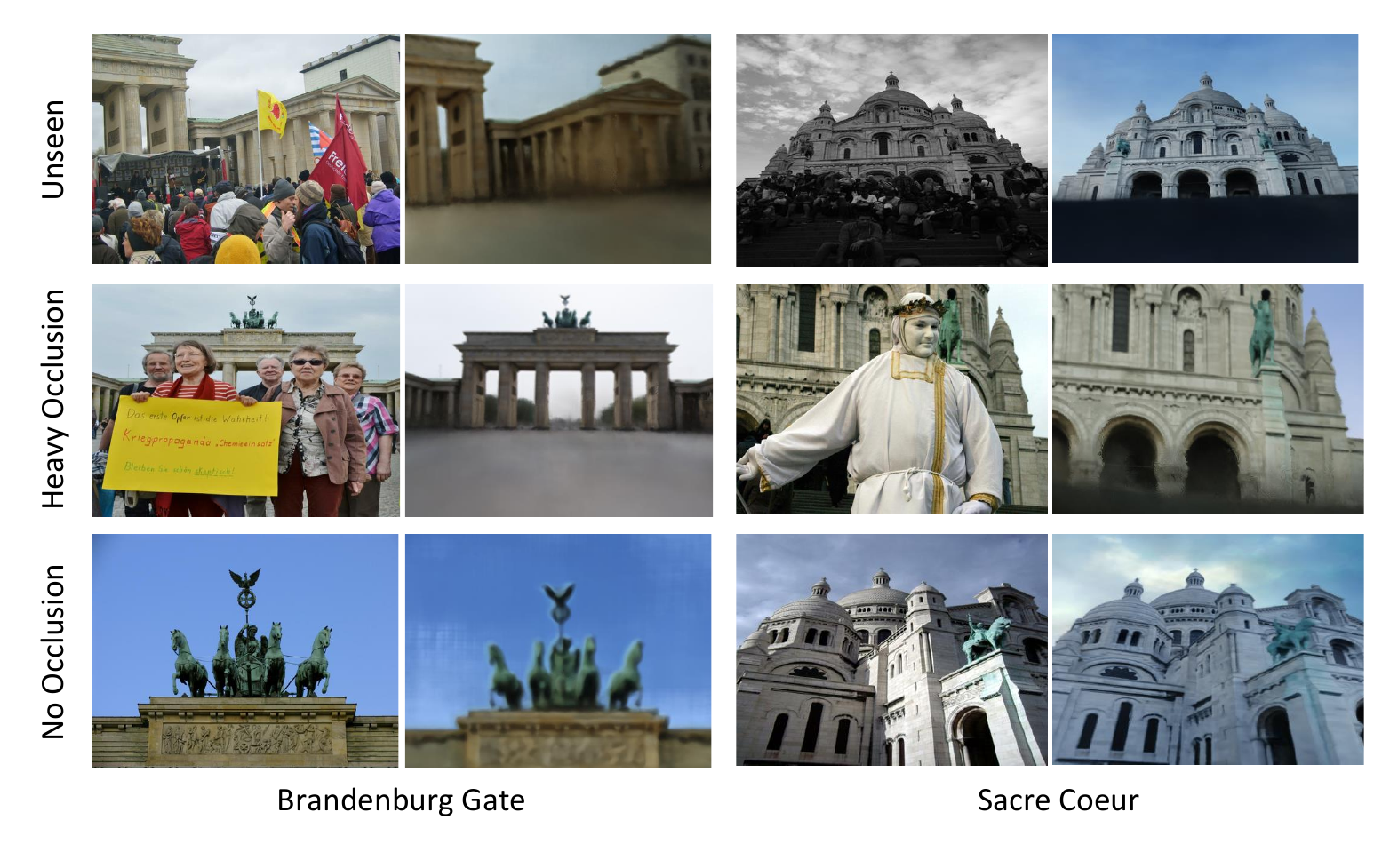}
\caption{Examples of generalization performance in unseen and severe conditions of IE-NeRF.}
\label{fig: generalization ability}
\end{figure*}

\paragraph{Genaration Ability}
\chadded[id=R2]{We conduct experiments on unseen appearances, to assess the generalization ability of IE-NeRF in the wild, as shown in \figurename~\ref{fig: generalization ability}, while the performance is not as well as regular, the model still demonstrate a reasonable level of accuracy, indicating its ability to generalize in the wild.}
\chadded[id=R2]{To further evaluate the model's performance under severe conditions, we test the model on heavily occluded images. The results showe that the model's performance does not degrade significantly under high occlusion. Conversely, when tested in environments without occlusions, the model, while not performing exceptionally well, still demonstrate overall robustness. Examples of these cases can be seen in \figurename~\ref{fig: generalization ability}, there is still a need for further optimization.}

\section{Limitations and Future Work}

Our \ours\ focuses on solving the novel view synthesis from the photographs in the wild by removing and inpainting the transient phenomena. However, this approach still struggles with small datasets and sparse inputs, as the transient mask lacks sufficient information to infer and guarantee the inpainting process. To improve further, an approach to learning the consistent appearance of the static scene with varying transient phenomena in the few-shot setting is necessary. Moreover, the camera parameters for each image, derived from structure-from-motion, exhibit some inaccuracies. As a result, our future work will involve concurrently refining the camera poses to enhance accuracy.

\section{Conclusion}
	
\label{sec:con}
This paper proposes a novel method to enhance the NeRF in the wild. Our approach improves traditional NeRF by integrating inpainting that helps eliminate occlusions and restore the static scene image, guided by the transient mask generated from the MLPs of the extending NeRF network. Results from both qualitative and quantitative experiments on Phototourism datasets demonstrate the effectiveness of our method in novel view synthesis, particularly under the challenge of transient phenomena. Additionally, we propose a new training strategy using frequency regularization with the transient mask factor in integrated positional encoding. Ablation studies further verify that this strategy facilitates faster inference and early separation of transient components during training. Currently, our proposed approach still encounters challenges on small datasets or under sparse inputs, as the transient mask lacks sufficient information to guide the inpainting process. For further optimization, we plan to explore an approach to learning the independent appearance of the static scene with varying transient phenomena under the few-shot setting.

\bibliographystyle{unsrt}
\bibliography{ref}

\end{document}